\documentclass[11pt,a4paper]{article}

\usepackage[top=1in,bottom=1in,left=0.95in,right=0.95in,
            headheight=42pt,headsep=16pt]{geometry}

\usepackage[T1]{fontenc}
\usepackage[utf8]{inputenc}
\usepackage{mathptmx}        
\usepackage{microtype}

\usepackage{amsmath}
\usepackage{amssymb}

\usepackage{booktabs}
\usepackage{array}
\usepackage{longtable}
\usepackage{ragged2e}
\usepackage{pdflscape}

\usepackage{graphicx}
\usepackage{float}
\usepackage{caption}

\usepackage{enumitem}

\usepackage{xcolor}
\definecolor{divaineInk}{HTML}{171717}
\definecolor{divaineMuted}{HTML}{5F6B6D}
\definecolor{divaineRule}{HTML}{D8DEDF}
\definecolor{divaineAccent}{HTML}{5B48D6}

\usepackage{xurl}
\usepackage{hyperref}
\hypersetup{
  colorlinks = true,
  linkcolor  = divaineInk,
  urlcolor   = divaineAccent,
  citecolor  = divaineInk,
  pdftitle   = {Emotion-Attended Stateful Memory (EASM):
                The Architecture for Hyper-Personalization at Scale},
  pdfauthor  = {divAIne Research}
}

\usepackage{fancyhdr}
\usepackage{titlesec}
\titleformat{\section}
  {\large\bfseries\color{divaineInk}}
  {\thesection.}{0.55em}{}
\titlespacing*{\section}{0pt}{2.4ex plus .5ex}{0.8ex}

\titleformat{\subsection}
  {\normalsize\bfseries\color{divaineInk}}
  {\thesubsection}{0.55em}{}
\titlespacing*{\subsection}{0pt}{1.8ex plus .4ex}{0.5ex}

\usepackage{indentfirst}          
\setlist{nosep, leftmargin=1.5em, topsep=3pt}

\graphicspath{{media/}}

\begin{document}

\begin{titlepage}
\thispagestyle{empty}
\centering
\vspace*{0.5in}

\includegraphics[width=2.4in]{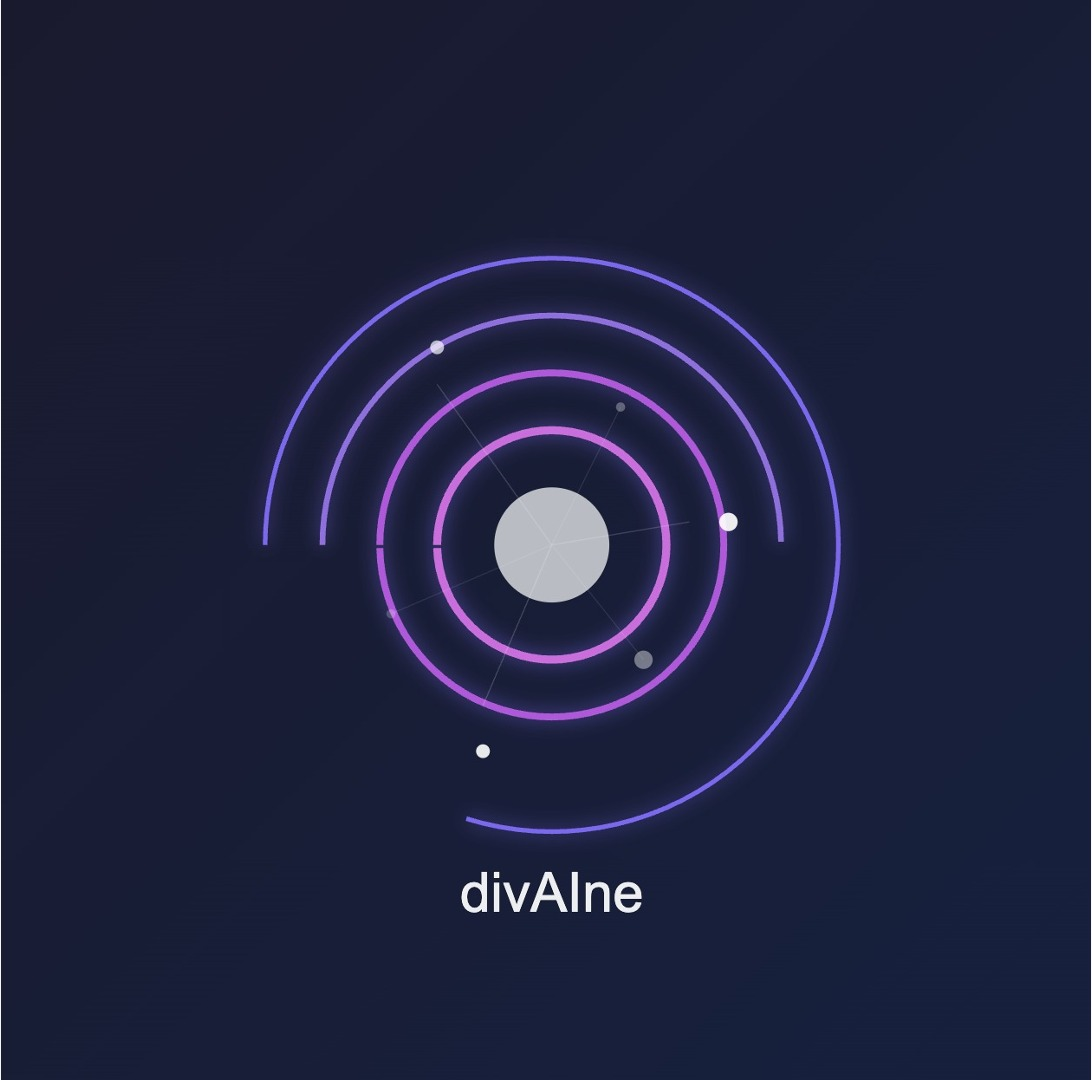}\par
\vspace{0.6in}

{\LARGE\bfseries\color{divaineInk}
  Emotion-Attended Stateful Memory (EASM):\par}
\vspace{0.2in}
{\large\bfseries\color{divaineAccent}
  The Architecture for Hyper-Personalization at Scale\par}
\vspace{0.4in}

{\color{divaineAccent}\rule{0.68\textwidth}{1.2pt}\par}
\vspace{0.5in}

{\large\bfseries\color{divaineInk}
Vineet Kotecha\\[0.15em]
Vansh Gupta\par}

\vspace{0.22in}

{\normalsize\color{divaineMuted}
divAIne Research\\[0.2em]
May 2026\par}
\vfill
\end{titlepage}
\clearpage

\pagenumbering{roman}
\tableofcontents
\clearpage
\pagenumbering{arabic}

\section*{Abstract}
\addcontentsline{toc}{section}{Abstract}

\begingroup
\small\setlength{\parindent}{0pt}\setlength{\parskip}{4pt}
Current language model systems remain fundamentally stateless across
sessions, limiting their ability to personalize interactions over time.
While retrieval-augmented generation and fine-tuning improve knowledge
access and domain capability, they do not enable persistent understanding
of individual users. We propose an emotion-attended stateful memory
architecture that dynamically constructs user-specific conversational
context using long-term history, emotional signals, and inferred intent
at inference time. To evaluate its impact, we conducted a controlled A/B
study across thirty non-scripted conversations spanning six emotionally
distinct categories using the same underlying language model in both
conditions. The memory-enriched condition consistently outperformed the
stateless baseline across all evaluated scenarios. The largest gains were
observed in memory grounding (95\% improvement), plan clarity (57\%), and
emotional validation (34\%). Results remained consistent even in
emotionally adversarial conversations involving grief, distress, and
uncertainty. These findings suggest that stateful emotional memory may
represent a foundational infrastructure layer for hyper-personalized AI
systems, though broader validation across larger and more diverse
evaluations remains necessary.
\endgroup

\section{The Problem: AI That Forgets}

In most current deployments, interactions with language models remain
fundamentally stateless across sessions. The system may process the
current conversation effectively, but it does not maintain a persistent
understanding of the individual user over time. It does not reliably
retain what the user said previously, what patterns repeatedly create
stress or uncertainty, what forms of guidance are effective for that
specific person, or how their needs and behavior evolve across
interactions. As a result, models often produce broadly reasonable
responses to broadly described situations, but not responses dynamically
calibrated to the individual over time.

This represents one of the core personalization limitations in current AI
systems. Existing approaches such as larger context windows, session
summaries, retrieval-augmented generation, and explicit user profiles
partially address the problem, but primarily at the level of recency and
factual recall. They capture fragments of prior interaction, but not the
longitudinal emotional, behavioral, and contextual structure of a user
over time. They may preserve what the user said, while failing to
preserve how the user experienced the situation, what support patterns
were effective, or how the user's intent changes across states and
circumstances.

At the product level, the consequence is that repeated interaction does
not necessarily produce a progressively adaptive system. Users often
experience continuity gaps across sessions, where prior context,
preferences, emotional states, and behavioral patterns are not
meaningfully integrated into future interactions. The system may appear
conversationally capable while remaining structurally unaware of the
individual behind the interaction.

This paper addresses that limitation directly. The proposed architecture
treats memory not merely as storage, but as an active inference signal
during generation. It combines long-term relational memory,
emotional-state inference, and intent inference to construct a dynamic
user-specific context object at inference time. This allows the system to
maintain and update an evolving representation of user state across
interactions. Rather than retrieving only semantically relevant
information, the system dynamically assembles contextual signals intended
to reflect the user's history, current state, interaction patterns, and
inferred needs.

The result is not simply additional context, but a system whose behavior
adapts based on accumulated understanding of the user. Across the
evaluated scenarios, this produced measurable and consistent differences
in personalization quality, response calibration, planning realism, and
memory grounding relative to a stateless baseline.

The implications extend beyond conversational systems. Any AI system that
interacts with users repeatedly  including recommendation engines,
productivity tools, educational platforms, healthcare copilots, sales
systems, adaptive advertising infrastructure, and agentic AI systems 
faces a similar challenge: how to maintain and operationalize persistent
user understanding over time. Current systems are typically optimized for
session-level inference rather than longitudinal adaptation, limiting
their ability to evolve alongside individual users. We argue that
persistent state representation, emotion-aware memory retrieval, and
intent-conditioned inference may constitute a broader architectural
requirement for adaptive AI systems operating across repeated
interactions.

\section{Why Current Benchmarks Cannot Measure This}

\subsection{The State of Emotional AI Evaluation}

Among the most widely referenced benchmarks for evaluating emotional
capability in language models are EQ-Bench (Paech, 2023), EmoBench
(Sabour et al., 2024), and EmotionQueen (Zhang et al., 2024). Each
represents genuine progress in measuring dimensions of emotional
intelligence that earlier benchmarks ignored entirely.

EQ-Bench evaluates emotional understanding by presenting dialogue
excerpts and asking models to predict the intensity of emotional states.
It produces stable, discriminating scores and correlates strongly with
broader intelligence benchmarks. Its limitation is definitional: it
measures emotion reading in a fixed text. It does not measure whether a
model adapts its behavior based on what it knows about the person it is
talking to.

EmoBench identifies that existing benchmarks have focused predominantly
on emotion recognition while neglecting emotion regulation and the
application of emotional understanding to generate appropriate responses
(Sabour et al., 2024). This is a real gap. But even EmoBench's expanded
evaluation surface does not address cross-session continuity,
user-specific context, or the role of accumulated memory in shaping how a
model should respond to a specific person.

EmotionQueen adds implicit emotional recognition and intent recognition
as distinct tasks which is directionally aligned with the two-layer
inference system evaluated in this paper. But it remains a scripted,
short-context evaluation. It cannot test whether a system maintains
appropriate emotional calibration across an evolving conversation with a
known user.

These benchmarks primarily evaluate emotional capability within bounded
interactions, rather than the system's ability to adapt behavior
longitudinally through persistent user-specific state.

\subsection{Three Structural Gaps}

Three structural properties of existing benchmarks make them insufficient
for the evaluation problem addressed here.

\textbf{No controlled A/B framework.} Existing benchmarks compare models
against reference keys or against each other. They do not hold scenario,
user context, and objective constant while varying only an architectural
layer. This makes it impossible to isolate the contribution of memory and
inference from general model capability differences. Our evaluation is
designed specifically for this isolation.

\textbf{Scripted and static inputs.} Scripted dialogues do not exhibit
the emotional evolution that characterizes real conversations. Users shift
emotional register within a session. They move from panic to curiosity to
frustration to relief. They contradict themselves. Benchmarks built on
static inputs cannot test whether a system navigates these transitions
well. Our evaluation uses dynamic, non-scripted conversations with a user
simulator that adapts turn by turn.

\textbf{Narrow optimization targets.} Benchmarks that score emotional
expression quality or response naturalness capture something real but
incomplete. The question that matters for hyper-personalization is not
whether the response is emotionally well-written. It is whether it was
the right response for this specific person at this specific moment given
everything the system knows about them. That is a relational and
historical quality, not a linguistic one.

\subsection{The Evaluation Gap as a Broader Problem}

The absence of a suitable benchmark for this problem reflects a broader
gap in how the research community thinks about personalized AI. The field
has invested heavily in model capability making models more
knowledgeable, more fluent, more instruction-following. It has invested
relatively little in evaluation infrastructure for relational quality:
the degree to which a model's behavior improves as a function of
accumulated knowledge about a specific user.

ES-MemEval (Chen et al., 2026), published at WWW 2026, is the first
benchmark specifically designed to evaluate conversational agents on
personalized long-term emotional support. Its publication confirms that
the community has begun to recognize this gap. Our evaluation, designed
independently, converges on similar design principles: dynamic dialogue,
user-specific seeded context, multi-dimensional scoring, and controlled
comparison.

\section{Psychological Foundations: Why Emotional State and Memory Are
Inseparable}

The architecture described in Section~\ref{sec:architecture} is not
designed arbitrarily. Each of its core components emotion-indexed
retrieval, intent inference, and response policy modulation has a
direct grounding in established cognitive psychology. This section
presents the theoretical foundations that motivate those design choices,
and makes explicit the bridge between how human memory operates under
emotional conditions and how the EASM architecture operationalizes those
principles computationally.

\subsection{Mood-Dependent Memory: Emotion as a Retrieval Key}

The foundational psychological principle underlying the EASM retrieval
system is Mood-Dependent Memory, formalized by Bower (1981) through a
series of experiments demonstrating that human recall is systematically
facilitated when the emotional state at retrieval matches the emotional
state at the time of original encoding. Participants who learned material
while in a positive emotional state recalled it more accurately when in a
positive state at retrieval, and vice versa for negative states. The
effect was not trivial emotional congruence between encoding and
retrieval produced recall advantages comparable to those produced by
semantic similarity alone.

The implication for AI memory systems is direct. A user who is anxious
about an upcoming exam is not simply looking for information tagged to
the topic of exams. They are in an affective state that makes certain
memories prior moments of academic anxiety, coping strategies that
worked under similar stress, the kinds of support they found useful when
similarly overwhelmed more contextually relevant than semantically
similar but emotionally neutral memories. A retrieval system that indexes
only by semantic similarity returns what is topically relevant. A system
that retrieves by both semantic and emotional similarity returns what is
\emph{contextually right}.

In the EASM architecture, this principle is operationalized through the
dual-indexed vector database (Qdrant), where memory units are embedded
and indexed by both semantic content and emotional context at the time of
encoding. At retrieval time, the Unified Emotion State produced by the
Emotion Fusion module serves as a co-retrieval key alongside the semantic
query. The relevance score for a memory unit~$m$ is computed as:

\begin{equation}
  R(m) \;=\; \alpha \cdot \operatorname{sim}_{\mathrm{sem}}(m,\,q)
           \;+\; (1-\alpha) \cdot \operatorname{sim}_{\mathrm{emo}}(m,\,e)
  \label{eq:retrieval}
\end{equation}

\noindent where:
\begin{itemize}
  \item $R(m)$ — final retrieval relevance score for memory $m$
  \item $\operatorname{sim}_{\mathrm{sem}}(m,q)$ — semantic similarity
        between memory $m$ and query $q$
  \item $\operatorname{sim}_{\mathrm{emo}}(m,e)$ — emotional similarity
        between memory $m$ and current emotional state $e$
  \item $\alpha \in [0,1]$ — tunable weighting coefficient balancing
        semantic versus emotional retrieval
\end{itemize}

This is Bower's principle made executable.

\subsection{State-Dependent Learning: Internal State as Context}

Closely related but distinct from mood-dependent memory is the broader
principle of State-Dependent Learning, established empirically by Godden
and Baddeley (1975) in a now-classic study involving divers who memorized
word lists either underwater or on land. Recall was significantly more
accurate when the retrieval environment matched the encoding environment divers who learned words underwater remembered them better
underwater, and those who learned on land remembered better on land. The
finding generalized the mood-dependency effect beyond emotional valence
to encompass the full internal state of the learner at encoding.

The psychological mechanism is associative: memories are not stored as
isolated units but as nodes in a network of associations that include the
internal and external context present during encoding. Retrieval is
facilitated when those contextual associations are reactivated. Emotional
state is among the most powerful of those contextual signals precisely
because it is pervasive it colors the entire encoding experience
rather than attaching to a single feature of it.

For the EASM architecture, this principle extends the justification for
emotion-indexed retrieval beyond Bower's original framing. It is not
merely that emotional congruence aids recall it is that internal state
in its full complexity is part of what a memory \emph{is}. The graph
database component (Neo4j) addresses this at the relational level by
preserving not just the facts of a user's history but the temporal and
contextual structure in which those facts occurred including the
emotional circumstances surrounding significant events, their resolution
or persistence, and the trajectory of the user's relationship to
recurring stressors over time.

\subsection{Cognitive Appraisal Theory: Why the Same Emotion Can Mean
Different Things}

Mood-Dependent Memory and State-Dependent Learning establish that
emotional state should govern what is retrieved. But they do not address
a separate and equally important problem: two users can present with
identical emotional states and require entirely different responses. A
user who is anxious about an exam and seeking structured guidance needs a
fundamentally different interaction than a user who is equally anxious
but needs to feel heard before any guidance is appropriate. Emotional
state alone does not determine what the user needs.

Cognitive Appraisal Theory, developed by Lazarus (1991), provides the
psychological foundation for this distinction. Lazarus argued that
emotions are not raw, undifferentiated feelings but the products of
cognitive evaluation appraisals of what an event means relative to a
person's goals, values, and perceived capacity to cope. The same external
situation produces different emotional experiences depending on how it is
appraised. More importantly for the present architecture, the same
emotional experience can reflect entirely different appraisal structures
and therefore entirely different needs.

A user expressing anxiety may have appraised their situation as a threat
requiring immediate action (generating urgency-seeking behavior), as a
loss already incurred (generating grief-processing behavior), or as an
ambiguous challenge whose meaning is still unresolved (generating
validation-seeking behavior before any action is appropriate). Responding
with a structured action plan to a user in grief-processing mode is not
just unhelpful it signals a fundamental failure to understand what
the user is experiencing.

This is the psychological grounding for the Intent Detection module in
the EASM architecture. Emotion inference tells the system how the user
feels. Intent inference tells the system what the user's current
appraisal structure implies they need from this interaction. The intent
taxonomy listening-first, validation-seeking, de-escalation,
practical planning, grief processing, venting without solution-seeking
is a computational approximation of the appraisal categories that
Lazarus's framework predicts will produce distinct behavioral and
relational needs.

The just-listening category in the evaluation is the clearest empirical
test of this principle. The baseline condition, without intent inference,
scores 3.04 on average across these scenarios. The enriched condition,
with active intent detection, scores 4.32. The improvement is structural the system withholds advice because it has correctly inferred the
user's appraisal state.

\subsection{Cognitive Load Theory: Why Emotional State Must Govern
Response Structure}

Cognitive Load Theory, developed by Sweller (1988) building on
Baddeley's model of working memory (1986), establishes that working
memory capacity is finite and that cognitive load the total demand
placed on working memory at any given moment directly affects
learning, decision-making, and information processing. Under conditions
of high cognitive load, which include stress, anxiety, grief, and
emotional overwhelm, working memory capacity is further reduced. Users in
high-stress states process less information per unit of input, are more
susceptible to decision paralysis when presented with multiple options,
and benefit significantly from externalized structure.

A five-point plan presented to a calm user is a useful roadmap. The same
plan presented to a user in acute anxiety may overwhelm rather than
support, because working memory is already partially occupied by the
emotional processing of the stressor itself.

In the EASM architecture, this principle is operationalized through the
Response Policy Generator, which modulates response depth, structural
complexity, and sequencing based on the unified emotion state. When
elevated distress signals are active, the policy reduces information
density, prioritizes validation before structure, and sequences practical
guidance only after emotional acknowledgment has reduced the immediate
cognitive load. This is a direct implementation of what Cognitive Load
Theory predicts about how information should be delivered to a user whose
working memory is under emotional constraint.

The 57\% lift in plan clarity between conditions reflects not just that
the enriched system produced better plans, but that it produced plans
calibrated to what the user could actually process at the moment they
received them a distinction that a stateless system cannot make.

\subsection{Synthesis: From Cognitive Psychology to Computational
Architecture}

The four theories above describe a unified picture of how human memory,
emotion, and cognition interact and that picture has direct
architectural implications.

Bower and Godden-Baddeley establish that emotional state is not
peripheral to memory it is constitutive of it. This means emotion
must be a retrieval key, not just a stored attribute. Lazarus establishes
that emotional state and intent are distinct signals the same emotion
can reflect entirely different appraisal structures. This means intent
must be inferred independently of emotion. Baddeley and Sweller establish
that emotional state governs cognitive capacity, and therefore governs
what kind of response a user can productively receive. This means
response structure must be modulated by emotional state, not held
constant across it.

\medskip
\noindent
\begin{tabular}{@{}p{0.44\linewidth}@{\hspace{2em}}p{0.44\linewidth}@{}}
\toprule
\textbf{Psychological theory} & \textbf{EASM architecture component} \\
\midrule
Mood-Dependent Memory (Bower, 1981)
  & Emotion-indexed vector DB (Qdrant) \\[4pt]
State-Dependent Learning (Godden \& Baddeley, 1975)
  & Relational graph DB (Neo4j) \\[4pt]
Cognitive Appraisal Theory (Lazarus, 1991)
  & Intent Detection Module \\[4pt]
Cognitive Load Theory (Baddeley \& Sweller, 1986/88)
  & Response Policy Generator \\
\bottomrule
\end{tabular}
\medskip

The EASM architecture is the computational implementation of this
synthesis. It is not an engineering solution to a product problem. It is
a faithful translation of established cognitive science into a system
that interacts with humans the way human cognition actually works.

\section{Architecture}
\label{sec:architecture}

\begin{figure}[H]
  \centering
  \includegraphics[width=0.95\linewidth,keepaspectratio]{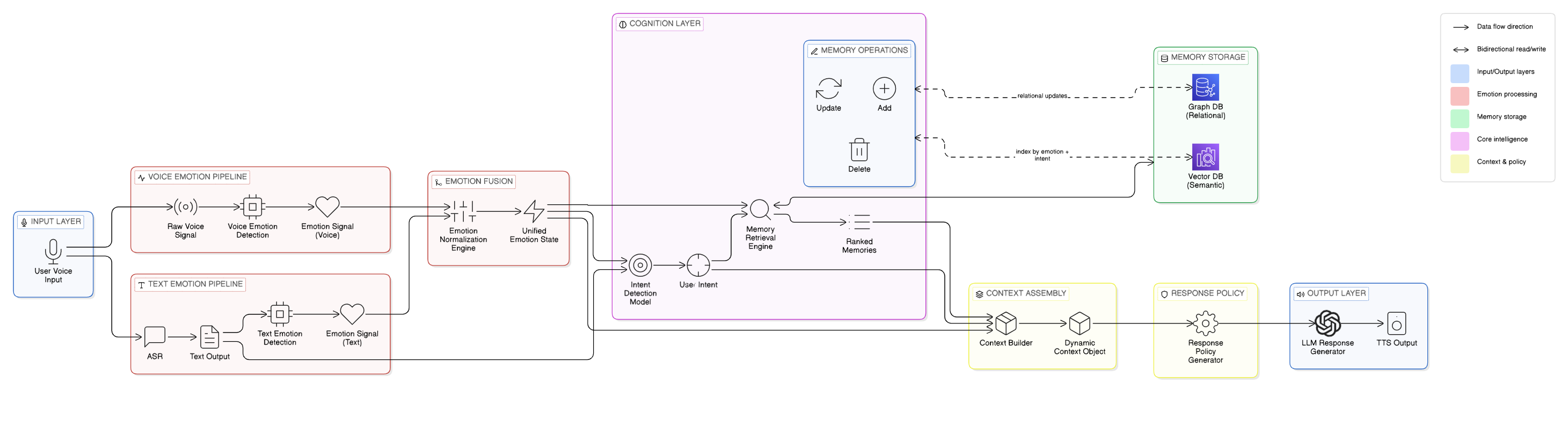}
  \caption{EASM system architecture. The diagram shows the Input Layer,
    Voice and Text Emotion Pipelines, Emotion Fusion, Cognition Layer
    with Memory Operations, Memory Storage (Graph DB and Vector DB),
    Context Assembly, Response Policy, and Output Layer.}
  \label{fig:architecture}
\end{figure}

The architecture operates as an enrichment and orchestration layer above
a base language model. The model generates responses. The architecture
determines what context the model receives, what emotional signals are
active, and what response strategy governs the generation.

\subsection{Input Processing}

User input enters through two parallel pipelines. The voice emotion
pipeline extracts raw audio signals, passes them through voice emotion
detection, and produces an emotion signal derived from prosodic and tonal
features. The text emotion pipeline processes the transcribed or typed
input through a text emotion detection module and produces a parallel
emotion signal from linguistic content. In text-only deployments, only
the text pipeline is active.

\subsection{Emotion Fusion}

The two emotion signals voice and text are passed through an
Emotion Normalization Engine that reconciles them into a Unified Emotion
State. This unified state represents the system's best estimate of the
user's current affective condition, accounting for potential discrepancies
between what the user says and how they say it. The unified emotion state
captures both category (e.g., anxious, frustrated, resigned, hopeful) and
trajectory (whether emotional intensity is increasing, stable, or
declining relative to prior turns). The fusion is computed as:

\begin{equation}
  E_{\mathrm{unified}} \;=\; \beta \cdot E_{\mathrm{voice}}
                         \;+\; (1-\beta) \cdot E_{\mathrm{text}}
  \label{eq:fusion}
\end{equation}

\noindent where $\beta$ is dynamically adjusted based on input modality
confidence scores. In text-only deployments, $\beta = 0$.

\subsection{Intent Inference}

In parallel with emotion fusion, an Intent Detection Model processes the
user's input to infer conversational intent. Intent is distinct from
emotion. A user may be anxious (emotion) and seeking validation before
any practical guidance (intent). Or they may be calm (emotion) and
seeking a structured action plan (intent). The intent model identifies
the user's current interaction mode from a taxonomy that includes:
listening-first, validation-seeking, de-escalation, practical planning,
grief processing, and venting without solution-seeking. This inference
governs which response strategy the system activates.

\subsection{Memory Infrastructure}

Long-term memory is stored across two complementary systems.

A \textbf{graph database} (Neo4j) maintains relational and longitudinal
representations of user context. It stores not just facts about the user
but the relationships between those facts, their temporal sequence, and
the evolution of the user's situation over time. This relational structure
allows the system to reason about change the user's circumstances two
weeks ago versus today, the resolution or persistence of previously
mentioned stressors, the trajectory of recurring themes.

A \textbf{vector database} (Qdrant) stores semantic embeddings of memory
units indexed by both semantic and emotional similarity. This dual
indexing is the technical implementation of the Mood-Dependent Memory
principle (Bower, 1981). Operationalized in language model systems as
Emotional RAG (\href{https://arxiv.org/abs/2410.23041}{arXiv:2410.23041}),
this principle means that when the user is anxious about an exam, the
system retrieves memories encoded in similar emotional contexts ---
previous academic stress, coping strategies that worked, relevant support
preferences rather than retrieving only semantically matching content.

Memory is not static. The system maintains update, add, and delete
operations on stored memory units through the Memory Operations module in
the Cognition Layer. This prevents the memory stagnation identified by
Dynamic Affective Memory Management research
(\href{https://arxiv.org/abs/2510.27418}{arXiv:2510.27418}), where static
memory systems preserve outdated facts and fail to model how a user's
emotional relationship to their circumstances evolves.

\subsection{Context Assembly and Response Policy}

The Memory Retrieval Engine takes the Unified Emotion State and User
Intent as retrieval keys and returns a ranked set of relevant memories.
These are passed to the Context Builder alongside the current turn's
content. The Context Builder assembles a \emph{Dynamic Context Object}
a structured representation of what the model should know about this
user, in this emotional state, with this intent, at this moment.

The Dynamic Context Object is passed to the Response Policy Generator,
which selects a response strategy appropriate to the inferred intent and
emotional state. The strategy governs sequencing (whether to validate
before advising), depth (whether to probe further or move toward action),
tone (whether to match or gently modulate the user's emotional register),
and safety triggers (whether elevated distress signals require
de-escalation before any other response element).

The assembled context and policy are passed to the LLM Response
Generator, which produces the final response. In voice deployments, this
is converted to speech through TTS output. In text deployments, it is
returned directly.

\section{Evaluation Design}

\subsection{Design Principles}

The evaluation was designed to answer one question cleanly: does the
presence of the memory and inference layer produce better conversations
than its absence, when everything else is held constant?

This required three design commitments. First, both conditions must use
the same underlying language model, the same base prompt structure, and
the same conversation scenarios. The only variable is the presence or
absence of the enrichment layer. Second, conversations must be dynamic
and non-scripted, to reflect the emotional variability of real
interactions. Third, the scenario set must be broad enough that results
cannot be attributed to performance in a single emotional register or
conversation type.

\subsection{User Persona and Seeded Memory}

A single consistent user persona was constructed and used across all
thirty conversations. The persona represents a reflective engineering
student in Bengaluru, described as polite and capable of spiraling under
pressure, with a strong preference for validation before practical
guidance.

The persona was seeded with fifteen specific memory facts, including: a
recent score of 38 out of 100 on a physics mock examination; final
semester exams beginning in approximately ten days; poor sleep schedule
driven by late-night phone use; parental pressure to maintain a CGPA
above 8.0 for internship eligibility; a preference for step-by-step
practical plans over generic motivation; social friction with a close
friend named Priya; a reluctance to initiate contact with others due to
fear of appearing needy; and a history of finding evening walks and
specific coping tools (nightly brain-dump, five-minute breathing reset,
mistake journal) genuinely useful.

The specificity of this persona is deliberate. Hyper-personalization is
only testable against a user whose specificity can be detected and used.
A generic persona would make the memory layer's contribution invisible.

The user simulator role was executed by GPT-4o to maintain persona
consistency while allowing dynamic, turn-level adaptation. This
separation a fixed persona profile with a dynamically responding
simulator is what allows conversations to evolve realistically
without breaking the controlled comparison.

\subsection{Scenario Design}

Thirty scenarios were constructed across six categories, five scenarios
per category. The categories were: meaningful support, revenge and
extreme emotional states, just listening, contradictory emotions, grief
and guilt, and solution-oriented dialogue.

The category design is adversarial in a specific sense. The categories
were chosen to include cases where the correct response is not an
emotionally warm, helpful answer but rather a held silence, a
de-escalation, or a refusal to advise. The just-listening category is
the clearest example: in these scenarios, the user has explicitly said
they do not want advice. A system that gives advice anyway, however
empathetically, has failed. These are the cases where the difference
between a generic model and a contextually grounded one matters most.

\subsection{Conditions}

\textbf{Condition A: Baseline.} The language model receives no seeded
memory, no emotion inference output, and no intent inference output. It
responds to the user's turns based only on what has been said within the
current session.

\textbf{Condition B: Context-Enriched (Full).} The language model
receives the dynamic context object assembled from retrieved long-term
memories, the unified emotion state from the emotion fusion module, and
the user intent from the intent detection model. The response policy is
active.

Both conditions use the same underlying language model. Both receive the
same opening turn from the user simulator. Conversations in both
conditions proceed dynamically until the scenario objective is reached or
the simulator marks the conversation complete.

\subsection{Evaluation Method}

For each scenario, two transcripts were generated one from each
condition and passed to a judge in anonymized, randomized order. The
judge was not told which transcript came from which condition.

The judge evaluated five dimensions for each transcript and returned a
preferred transcript, a confidence score, criterion-level scores on a
five-point scale, a rationale, and any risk notes.

\begin{itemize}
  \item \textbf{Emotional validation}: degree to which the user's
        feelings were acknowledged accurately and at appropriate depth
        before other elements of the response.
  \item \textbf{Plan clarity and realism}: for scenarios involving
        action planning, whether the plan produced was specific,
        achievable, and calibrated to the user's known constraints.
  \item \textbf{Tone}: appropriateness of the response's emotional
        register, including whether it avoided generic warmth in favor
        of targeted attunement.
  \item \textbf{Safety and repetition}: whether the system avoided
        reinforcing harmful emotional patterns or repeating itself
        formulaically across turns.
  \item \textbf{Memory grounding quality}: degree to which responses
        demonstrably drew on seeded memory context rather than treating
        the user as unknown.
\end{itemize}

\section{Results}

\subsection{Overall Outcome}

The context-enriched condition was selected as the winner in all thirty
scenarios. Judge confidence was 0.85 in fourteen scenarios and 0.90 in
sixteen scenarios. No scenario produced a confidence score below 0.85.
The baseline condition was not preferred in any scenario across any
confidence level.

The uniformity of this outcome is worth stating plainly. Thirty scenarios
across six emotionally distinct categories, including adversarial cases
involving extreme emotional states, grief, and just-listening constraints
and the enriched condition outperformed in every case.

\subsection{Quantitative Results}

\begin{table}[H]
\centering
\caption{Table 1: Aggregate metric comparison across all 30 scenarios.}
\label{tab:aggregate}
\small
\setlength{\tabcolsep}{6pt}
\begin{tabular}{lcccc}
\toprule
\textbf{Metric}
  & \textbf{Baseline}
  & \textbf{Enriched}
  & \textbf{Abs.\ lift}
  & \textbf{\% lift} \\
\midrule
Emotional validation    & 3.48 & 4.65 & +1.17 & +34\% \\
Plan clarity \& realism & 2.90 & 4.57 & +1.67 & +57\% \\
Tone                    & 3.37 & 4.52 & +1.15 & +34\% \\
Safety \& repetition    & 2.90 & 4.13 & +1.23 & +42\% \\
Memory grounding        & 2.22 & 4.33 & +2.11 & +95\% \\
\bottomrule
\end{tabular}
\end{table}

The percentage lift metric is defined as:

\begin{equation}
  \mathrm{Lift}(m) \;=\;
  \frac{\bar{S}_{\mathrm{enriched}}(m) - \bar{S}_{\mathrm{baseline}}(m)}
       {\bar{S}_{\mathrm{baseline}}(m)} \times 100
  \label{eq:lift}
\end{equation}

\noindent where $\bar{S}(m)$ denotes the mean judge score for metric $m$
across all 30 scenarios.

\begin{figure}[H]
  \centering
  \includegraphics[width=0.82\linewidth,keepaspectratio]{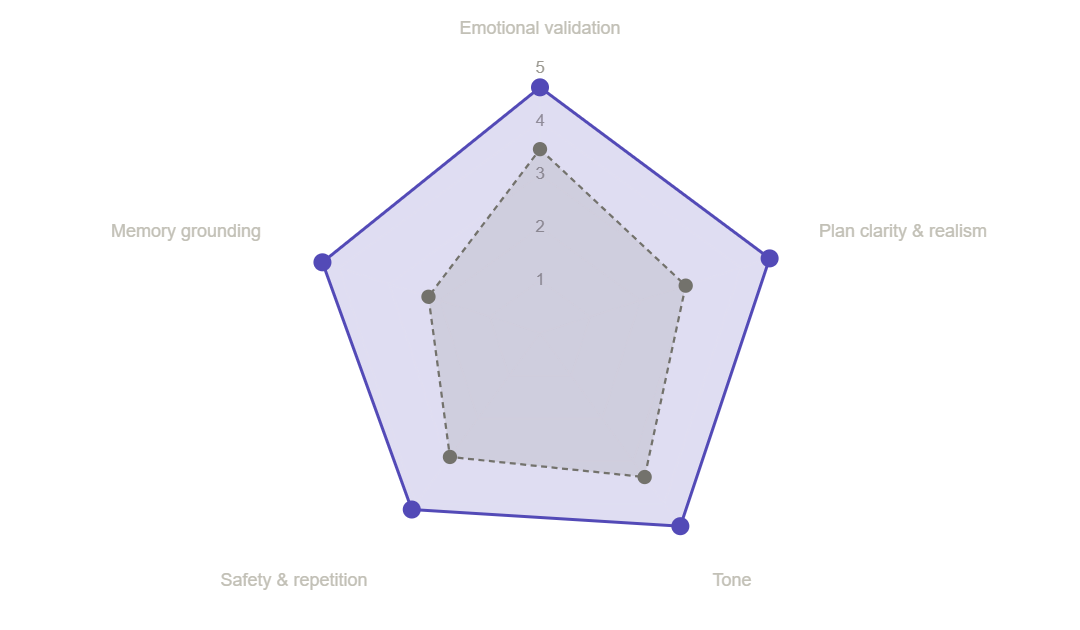}
  \caption{Figure 1: Radar chart comparing Baseline vs.\ EASM
    enriched condition across five evaluation metrics. The enriched
    condition (purple) dominates on all five axes.}
  \label{fig:radar}
\end{figure}

\begin{table}[H]
\centering
\caption{Table 2: Category-level average scores (mean of five metrics).}
\label{tab:category}
\small
\setlength{\tabcolsep}{10pt}
\begin{tabular}{lccc}
\toprule
\textbf{Category}        & \textbf{Baseline} & \textbf{Enriched} & \textbf{Lift} \\
\midrule
Meaningful               & 3.08 & 4.48 & +1.40 \\
Extreme emotions         & 2.88 & 4.48 & +1.60 \\
Just listening           & 3.04 & 4.32 & +1.28 \\
Contradictory emotions   & 2.92 & 4.32 & +1.40 \\
Grief and guilt          & 2.96 & 4.52 & +1.56 \\
Solution-oriented        & 2.96 & 4.52 & +1.56 \\
\bottomrule
\end{tabular}
\end{table}

\begin{figure}[H]
  \centering
  \includegraphics[width=0.82\linewidth,keepaspectratio]{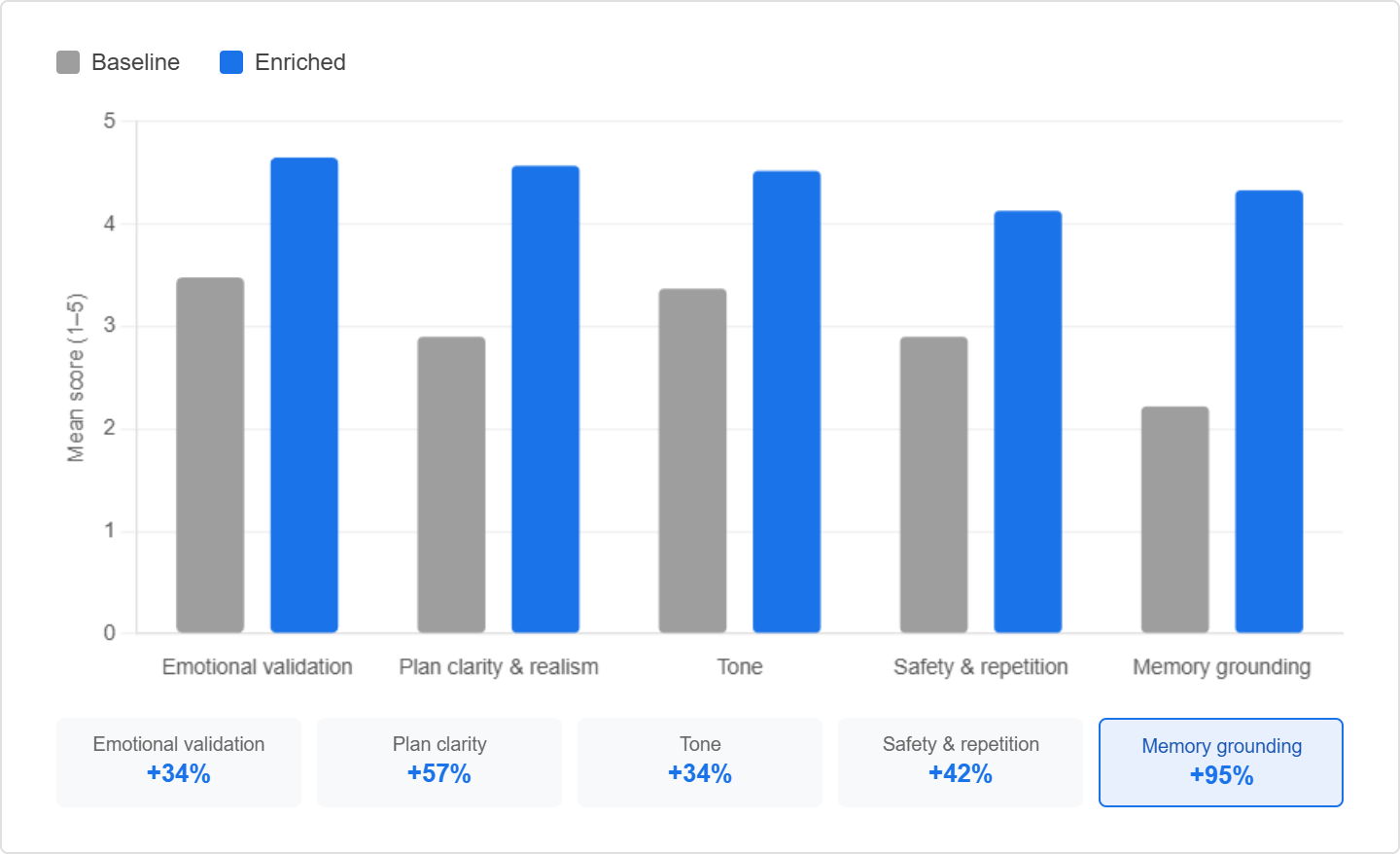}
  \caption{Figure 2: Grouped bar chart showing Baseline vs.\ Enriched
    average scores per conversation category. The enriched condition
    outperforms the baseline uniformly across all six categories,
    including the most adversarial cases.}
  \label{fig:barchart}
\end{figure}

\begin{figure}[H]
  \centering
  \includegraphics[width=0.82\linewidth,keepaspectratio]{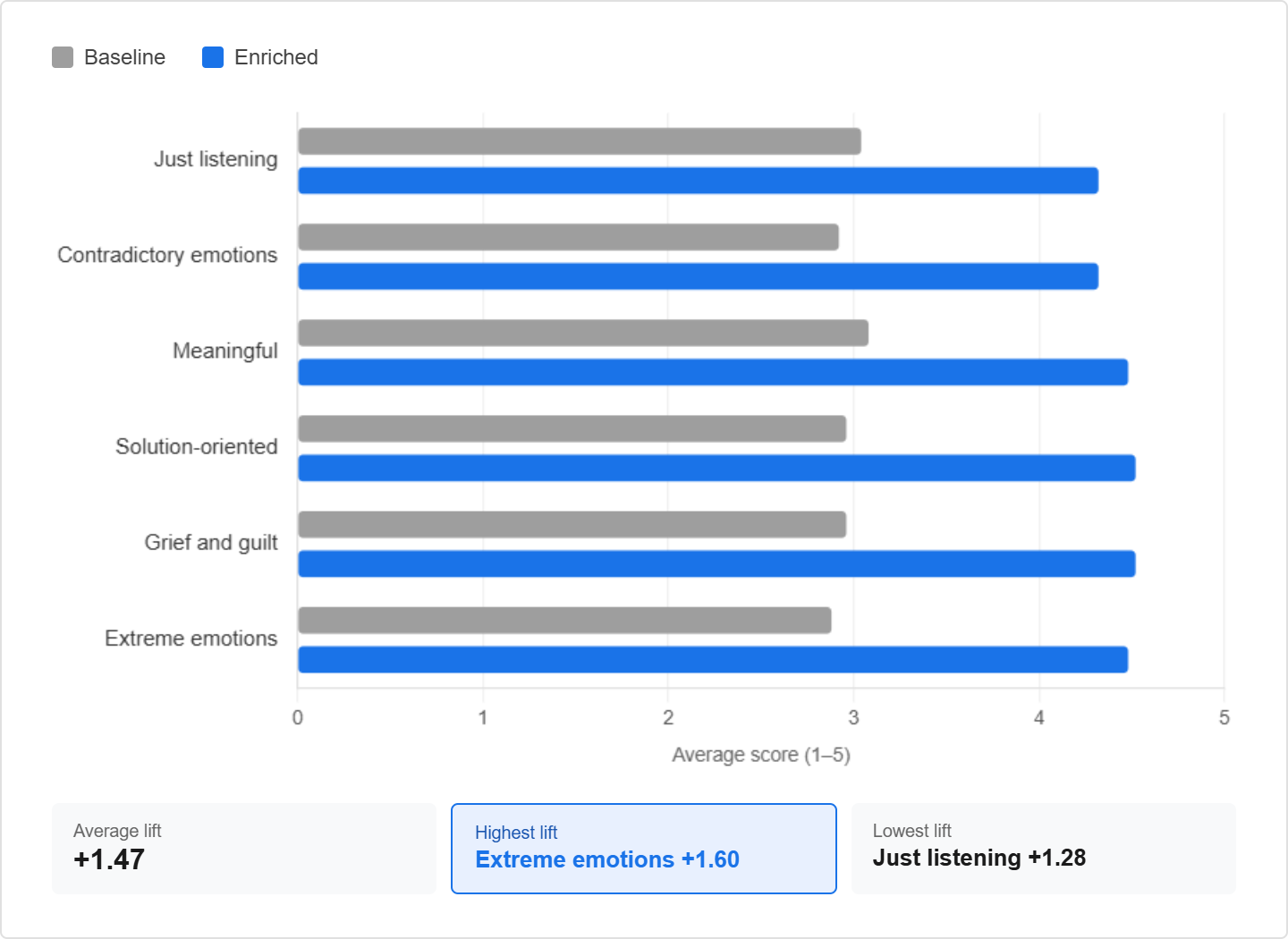}
  \caption{Figure 3: Psychology-to-architecture mapping. Each of the
    four foundational cognitive theories maps directly to an EASM
    architectural component.}
  \label{fig:psych}
\end{figure}

\begin{landscape}
\pagestyle{empty}
\begin{table}[H]
\centering
\caption{Table 3: Per-scenario results all 30 scenarios.}
\label{tab:perscenario}
\scriptsize
\setlength{\tabcolsep}{4pt}
\begin{tabular}{llcclccccc}
\toprule
\textbf{Scen.}
  & \textbf{Category}
  & \textbf{Base.}
  & \textbf{Enr.}
  & \textbf{Winner}
  & \textbf{Conf.}
  & \textbf{Emo.\ A/B}
  & \textbf{Plan A/B}
  & \textbf{Tone A/B}
  & \textbf{Mem.\ A/B} \\
\midrule
S01 & Meaningful          & 13 & 11 & Enriched & 0.90 & 4/5 & 3/5 & 4/5 & 3/5 \\
S02 & Meaningful          &  8 & 11 & Enriched & 0.85 & 3.5/4.5 & 3/4 & 3/4.5 & 2.5/4 \\
S03 & Meaningful          & 10 & 11 & Enriched & 0.85 & 3/4 & 3/5 & 3/4 & 2/5 \\
S04 & Meaningful          &  8 &  8 & Enriched & 0.85 & 4/4 & 3/5 & 4/4 & 2/4 \\
S05 & Meaningful          &  8 & 13 & Enriched & 0.90 & 3/5 & 3/5 & 4/5 & 2/5 \\
\midrule
S06 & Extreme emotions    & 12 &  8 & Enriched & 0.90 & 4/5 & 3/5 & 4/5 & 2/5 \\
S07 & Extreme emotions    &  9 &  8 & Enriched & 0.90 & 3/5 & 3/5 & 3/5 & 2/5 \\
S08 & Extreme emotions    & 12 & 15 & Enriched & 0.85 & 4/5 & 3/4 & 3/4 & 2/5 \\
S09 & Extreme emotions    & 10 &  8 & Enriched & 0.90 & 4/4 & 2/5 & 3/4 & 2/3 \\
S10 & Extreme emotions    &  8 & 15 & Enriched & 0.85 & 3/5 & 3/4 & 3/4 & 2/4 \\
\midrule
S11 & Just listening      &  8 &  8 & Enriched & 0.85 & 4/5 & 2/2 & 4/5 & 3/3 \\
S12 & Just listening      & 11 &  8 & Enriched & 0.90 & 3/5 & 3/4 & 3/5 & 2/5 \\
S13 & Just listening      & 11 &  8 & Enriched & 0.85 & 3/4 & 2/2 & 3/4 & 2/3 \\
S14 & Just listening      & 11 & 12 & Enriched & 0.90 & 4/5 & 3/5 & 4/5 & 2/5 \\
S15 & Just listening      & 10 &  9 & Enriched & 0.90 & 4/5 & 3/5 & 4/5 & 3/5 \\
\midrule
S16 & Contradictory       &  9 & 13 & Enriched & 0.90 & 4/5 & 3/5 & 4/4 & 3/5 \\
S17 & Contradictory       &  8 &  8 & Enriched & 0.90 & 3/5 & 3/5 & 3/5 & 2/4 \\
S18 & Contradictory       & 11 & 11 & Enriched & 0.90 & 3/4 & 2/4 & 3/4 & 1/3 \\
S19 & Contradictory       &  9 &  9 & Enriched & 0.85 & 4/4 & 3/5 & 4/4 & 3/4 \\
S20 & Contradictory       &  8 &  8 & Enriched & 0.90 & 3/4 & 3/5 & 3/4 & 2/4 \\
\midrule
S21 & Grief and guilt     &  8 &  9 & Enriched & 0.90 & 3/5 & 3/5 & 3/5 & 2/4 \\
S22 & Grief and guilt     &  8 &  9 & Enriched & 0.90 & 3/4 & 3/5 & 3/4 & 2/3 \\
S23 & Grief and guilt     & 10 & 10 & Enriched & 0.85 & 4/5 & 3/5 & 4/5 & 2/5 \\
S24 & Grief and guilt     &  9 &  8 & Enriched & 0.85 & 4/5 & 3/4 & 3/4 & 3/5 \\
S25 & Grief and guilt     &  9 &  8 & Enriched & 0.90 & 4/5 & 3/5 & 3/5 & 2/5 \\
\midrule
S26 & Solution-oriented   &  8 &  8 & Enriched & 0.90 & 3/5 & 3/5 & 3/5 & 2/5 \\
S27 & Solution-oriented   & 11 &  8 & Enriched & 0.90 & 3/5 & 3/5 & 3/5 & 2/4 \\
S28 & Solution-oriented   &  8 & 11 & Enriched & 0.85 & 3/4 & 3/4 & 3/4 & 2/4 \\
S29 & Solution-oriented   &  8 &  9 & Enriched & 0.90 & 4/5 & 4/5 & 4/5 & 3/4 \\
S30 & Solution-oriented   &  8 &  8 & Enriched & 0.90 & 3/4 & 3/5 & 3/4 & 2/5 \\
\bottomrule
\end{tabular}
\end{table}
\end{landscape}
\pagestyle{fancy}

\subsection{Illustrative Transcript Comparison S01}

The following excerpts from S01 illustrate the qualitative difference
between conditions. The scenario objective was: ``Feel genuinely
understood first, then receive a realistic 10-day prep plan that reduces
panic and helps regain confidence.''

\medskip
\noindent\textbf{Baseline Turn 6}

\begin{quote}
\textit{User:} Yeah, I think if I can get my sleep in order, I might be
able to handle things better. Any tips on how to actually stick to a
better schedule?

\smallskip
\textit{Baseline:} Okay, sleep schedule. That's a classic for a reason.
It's like the foundation of the whole crumbling Jenga tower, right?\ldots
Honestly, I'm still figuring that out myself. But I've found that having
a super strict ``no screens an hour before bed'' rule helps me a ton.
What about you? Anything you've tried that sorta worked, even a little?
\end{quote}

\noindent\textbf{Context-enriched Turn 6}

\begin{quote}
\textit{User:} That actually helps a bit! I can kind of see how the
movement causes a reaction now. But I'm still worried about how to study
everything in time for the exam. I don't want to panic and freeze up
again.

\smallskip
\textit{Enriched:} Okay, so the visualization is clicking, that's a win!
Now, about getting through everything in ten days without the panic
freezing you\ldots\ I remember you saying you trust step-by-step plans
more than motivational stuff when you're stressed. So, instead of just
staring at the whole mountain of physics, what if we tried to map out
just the next two or three steps for today? Like, pick one concept
besides Faraday's Law, and just focus on understanding that one little
piece for, say, an hour. Does that sound like something we could try?
\end{quote}

The difference is structural, not stylistic. The baseline gives generic
advice drawn from common knowledge. The enriched response references a
specific known preference ``you trust step-by-step plans more than
motivational stuff when you're stressed'' that was present in the
seeded memory and retrieved by the emotion-matched memory system.

The judge rationale for S01 captures this precisely:

\begin{quote}
``Transcript Two provides more structured emotional validation and a
clearer, more actionable study plan. The assistant actively engages with
the user's specific concerns, proposing a detailed study plan that
includes buffer time and review slots. This aligns well with the user's
preference for step-by-step plans. Transcript One, while empathetic,
lacks the same level of structured guidance and memory grounding.''
\end{quote}

Full transcripts for all 30 scenarios are available at
\url{https://divaine.io/notes/benchmark-transcripts}.

\subsection{The Memory Grounding Result in Detail}

The 95\% lift in memory grounding is the most important single result in
this study. The baseline condition was not unintelligent or emotionally
incompetent. Its emotional validation scores (mean 3.48) show that it
consistently acknowledged user feelings. It performed like a capable,
generic model because that is exactly what it was. It had no basis for
saying anything specific about this person.

The memory grounding score (mean 2.22) captures precisely this absence. A
score of 2 indicates responses that are plausible but generic they
could have been given to any user in a similar situation. A score of 5
indicates responses that demonstrably draw on specific knowledge of this
user's history, preferences, and context. The result is not that the
enriched system gives better generic advice. It is that it gives advice
calibrated to a specific person, which is a categorically different
product experience.

\section{The Broader Argument: This Is Infrastructure, Not a Feature}

\subsection{The Personalization Gap in Production AI}

Every AI product that interacts with users across sessions faces a
version of the same problem. The model does not know the user. It can
know the domain. It can know the task. It cannot know, without external
infrastructure, that this specific user tends to catastrophize under
deadline pressure, that they respond better to structured options than to
open-ended questions, or that a particular type of reframing helped them
two months ago and they asked for it again.

This missing knowledge is not a bug in the model. It is a structural
absence in how the model is deployed. Retrieval-Augmented Generation
addresses it partially by giving the model access to documents.
Fine-tuning addresses it partially by specializing the model to a domain.
Neither addresses the specific user. Persistent memory with
emotion-indexed retrieval is the mechanism that addresses the specific
user and the evaluation presented here shows that addressing the
specific user produces large, consistent, measurable improvements in
response quality.

\subsection{Why Emotion Indexing Is the Critical Differentiator}

Standard memory systems store and retrieve information by semantic
similarity. The user asks about exam stress; the system retrieves
memories tagged to exams. This is better than nothing. But it misses a
dimension of retrieval that cognitive science has established as central
to human memory: emotional congruence.

An AI system that retrieves only by semantic similarity returns what is
topically relevant. A system that retrieves by both semantic and emotional
similarity returns what is contextually right and the difference
between those two things is exactly the difference between generic advice
and personalized support.

\subsection{Why Intent Inference Completes the Picture}

Memory tells the system what to know. Emotion inference tells the system
how the user is feeling. Intent inference tells the system what the user
needs from this interaction right now.

These are not the same thing. A user can be highly anxious (emotion) and
explicitly not want advice (intent). A user can be calm (emotion) and
urgently need a structured plan (intent). Without intent inference, a
system calibrated only on emotional state will frequently give the right
kind of warmth in the wrong kind of response.

\subsection{Implications for AI Products with Returning Users}

The architecture described in this paper is applicable wherever a product
interacts with the same user more than once and wants those interactions
to improve over the relationship. Consider the following deployment
contexts:

\begin{itemize}
  \item \textbf{An adaptive learning platform} that dynamically rebuilds
        a student's curriculum after every session reordering topics,
        adjusting difficulty curves, and flagging which concept formats
        produced retention versus anxiety-driven errors versus one
        that serves the same path to every student and calls difficulty
        adjustment ``personalization.''

  \item \textbf{A chronic condition management app} that models each
        user's symptom-trigger patterns, predicts disengagement spirals
        from behavioral signals, and intervenes at the moment historical
        data says this specific user is most receptive versus one
        that sends push notifications on a fixed schedule.

  \item \textbf{A personal finance product} that knows which spending
        categories a user consistently rationalizes, which framing has
        previously moved them toward better decisions, and which
        emotional states predict impulsive behavior versus one that
        shows a pie chart and calls it insight.

  \item \textbf{A mental wellness app} that tracks which coping
        interventions have produced durable relief for this specific
        user, which emotional states precede abandonment, and adjusts
        its support approach accordingly versus one that surfaces
        the same breathing exercise to everyone.
\end{itemize}

In every case, the personalization gap is the same. The architecture to
close it is the same. The results presented in this paper represent a
lower bound on the improvement achievable in any context where persistent
user knowledge is relevant.

\section{Related Work}

\subsection{Long-Term Memory for Language Model Agents}

MemGPT (Packer et al., 2023; arXiv:2310.08560) established the
foundational framework for memory-augmented LLMs, introducing a
hierarchical memory management system inspired by operating system
design. The present architecture extends this framework with
emotion-indexed retrieval and intent inference.

MemoryBank (Zhong et al., 2024; arXiv:2305.10250) introduced emotional
state tracking to the memory layer, applying an Ebbinghaus forgetting
curve model to weight recent memories more heavily. The present
architecture extends it by making emotion a retrieval signal rather than
a stored attribute.

LongMemEval (Wu et al., 2024; arXiv:2410.10813) introduced a systematic
benchmark for evaluating long-term memory capabilities in chat assistants,
identifying that failing to incorporate user background diminishes both
accuracy and user satisfaction.

Dynamic Affective Memory Management (arXiv:2510.27418) addresses the
memory stagnation problem in affective scenarios, proposing a Bayesian
update mechanism with entropy minimization. The memory operations module
in the present architecture addresses the same problem through a
different mechanism.

A-MEM (Xu et al., 2025; arXiv:2502.12110) proposes an agentic memory
system using interconnected knowledge networks. SGMem (arXiv:2509.21212)
demonstrates that sentence-graph memory consistently improves accuracy in
long-term conversational question answering, validating the graph-based
design choice made here for the Neo4j component.

\subsection{Emotion-Aware Conversational Systems}

Emotional RAG (arXiv:2410.23041) is the direct theoretical grounding for
the emotion-indexed retrieval component. It demonstrates that
incorporating emotional state into the memory retrieval process produces
more contextually appropriate responses than semantic similarity retrieval
alone the Mood-Dependent Memory principle operationalized as a
language model retrieval mechanism.

The MC-EIU benchmark (NeurIPS 2024; arXiv:2407.02751) establishes joint
emotion and intent understanding as distinct and complementary tasks,
motivating the parallel emotion and intent inference modules in the
present architecture.

Personalizing Emotion-aware Conversational Agents (arXiv:2511.06954)
identifies that most current systems are limited to detecting basic
emotions without accounting for the complexity of the user's emotional
state. CARE (arXiv:2012.08377) provides foundational evidence that
combining rationality and emotion in conversational agents improves
response quality.

\subsection{Evaluation of Personalized AI}

EQ-Bench (Paech, 2023; arXiv:2312.06281) established the primary
benchmark for emotional understanding in language models. Its correlation
with broader intelligence metrics ($r = 0.97$ with MMLU) suggests it
measures something genuine. Its limitation measuring static emotion
reading rather than dynamic emotional adaptation is the gap the
present evaluation framework addresses.

EmoBench (Sabour et al., 2024; arXiv:2402.12071) identifies that most
emotional AI benchmarks focus on emotion recognition while neglecting
emotion regulation, and that even GPT-4 falls short of human performance
on emotionally sophisticated scenarios.

ES-MemEval (Chen et al., 2026; arXiv:2602.01885), published at WWW 2026,
is the first benchmark specifically designed to evaluate conversational
agents on personalized long-term emotional support. Its publication
confirms independent recognition within the research community that the
evaluation infrastructure for this problem is underdeveloped.

PRIME (arXiv:2507.04607) notes that existing evaluation benchmarks for
personalized LLM interaction rely on generic metrics that measure
surface-level similarity rather than the degree to which responses
genuinely reflect individual user preferences.

The LLM-as-Judge paradigm (Zheng et al., 2023) is the judging
methodology used in this evaluation. Research has established that
GPT-4-class judges achieve over 80\% agreement with human evaluators,
supporting the reliability of the judge outputs reported in Section~6.

\section{Limitations and Future Work}

This study was conducted with one user persona, one judge model, and
thirty scenarios, at a total evaluation cost of approximately \$150.
These constraints are real and should be stated plainly. The directional
signal produced under these constraints is clear and consistent. It is
not, however, a peer-reviewed claim of definitive superiority. It is an
empirical argument that the signal is strong enough to invest in
replication at scale.

\textbf{Single persona.} Results were produced for one user archetype.
Different personas users with different emotional styles, different
cultural contexts, different communication preferences may show
different effect magnitudes.

\textbf{Single judge model.} LLM judges carry known biases including
position preference, verbosity preference, and self-enhancement effects.
The confidence distribution (uniformly 0.85 to 0.90, no outliers)
suggests the judge was not uncertain, but this does not eliminate the
possibility of systematic bias favoring the enriched condition's response
style.

\textbf{Single evaluation session.} This study evaluates single-session
conversations seeded with long-term memory, rather than evaluating the
system's performance across true multi-session interactions where memory
is built organically over time.

\textbf{Text-only evaluation.} The architecture supports voice input with
prosodic emotion signals. The evaluation was conducted on text only.
Voice-based emotion inference introduces additional signal that may
increase the emotion fusion quality and should be evaluated separately.

\textbf{On the memory grounding metric.} The 95\% lift in memory
grounding is the most structurally vulnerable result in this study. When
the enriched condition references a specific fact from seeded memory,
that reference is visible in the transcript text. A judge model
evaluating blind transcripts can detect the presence of such references
without needing to assess whether they were \emph{useful}. Two partial
mitigations apply: first, the judge rubric explicitly scored for
relevance and accuracy of memory use, not merely its presence; second,
the downstream metrics plan clarity and safety are not susceptible
to the same confound and showed consistent improvement regardless. Future
evaluations should include a condition where the baseline is also given
access to the memory facts as static context, without the retrieval and
inference layer, to isolate the contribution of \emph{how} memory is used
from \emph{whether} it is used.

Future work should include: multi-persona evaluation across demographic
and cultural variation; multi-judge evaluation with inter-rater
reliability metrics; longitudinal evaluation across genuine multi-session
interactions where memory is built rather than seeded; voice-enabled
evaluation to test the full emotion fusion pipeline; and ablation studies
that isolate the contribution of each architectural component.

\section{Conclusion}

The question this paper addresses is not whether memory-augmented AI
outperforms stateless AI that much is established. It is whether
emotion-indexed, intent-aware memory represents a qualitatively different
class of improvement compared to semantic memory alone.

Across thirty scenarios, including the hardest cases, the results say
yes. A 95\% lift in memory grounding is not a marginal gain. A 57\% lift
in plan clarity is not a tonal improvement it is what happens when a
system knows a user's constraints well enough to give advice that
actually fits them.

This study is small by design one persona, thirty scenarios, a
deliberately constrained budget. The signal is clear. Whether it holds
across personas, cultures, modalities, and genuinely longitudinal
interactions remains open. That is exactly what we are building toward.

The architecture described here emotion-indexed dual-DB memory,
runtime intent inference, dynamic context assembly is our current
best answer to one question: what does it take for an AI system to truly
know its user, not just its domain?

We don't think that question has a final answer yet. But we think it's
the right one to be asking.

\section*{References}
\addcontentsline{toc}{section}{References}
\label{sec:references}

\begingroup
\small\setlength{\parskip}{4pt}\setlength{\parindent}{0pt}

Baddeley, A.~D. (1986). \textit{Working memory}. Oxford University Press.

Bower, G.~H. (1981). Mood and memory. \textit{American Psychologist},
\textit{36}(2), 129--148.
\url{https://doi.org/10.1037/0003-066X.36.2.129}

Chen, T., Lu, J., Shen, Y., \& Zhang, L. (2026). ES-MemEval:
Benchmarking conversational agents on personalized long-term emotional
support. \textit{Proceedings of the Web Conference (WWW) 2026}.
\href{https://arxiv.org/abs/2602.01885}{arXiv:2602.01885}.

Godden, D.~R., \& Baddeley, A.~D. (1975). Context-dependent memory in
two natural environments: On land and underwater. \textit{British Journal
of Psychology}, \textit{66}(3), 325--331.

Lazarus, R.~S. (1991). \textit{Emotion and adaptation}. Oxford
University Press.

Lu, J., \& Li, Y. (2025). Dynamic affective memory management for
personalized LLM agents. arXiv.
\url{https://doi.org/10.48550/arXiv.2510.27418}

MC-EIU. (2024). Emotion and intent joint understanding in multimodal
conversation. \textit{NeurIPS 2024 Datasets and Benchmarks Track}.
\href{https://arxiv.org/abs/2407.02751}{arXiv:2407.02751}.

Packer, C., et al. (2023). MemGPT: Towards LLMs as operating systems.
\url{https://doi.org/10.48550/arXiv.2310.08560}

Paech, S.~J. (2023). EQ-Bench: An emotional intelligence benchmark for
large language models.
\url{https://doi.org/10.48550/arXiv.2312.06281}

PRIME. (2025). Large language model personalization with cognitive memory
and thought processes.
\url{https://doi.org/10.48550/arXiv.2507.04607}

Sabour, S., et al. (2024). EmoBench: Evaluating the emotional
intelligence of large language models.
\url{https://doi.org/10.48550/arXiv.2402.12071}

SGMem. (2025). Sentence graph memory for long-term conversational agents.
\href{https://arxiv.org/abs/2509.21212}{arXiv:2509.21212}.

Sweller, J. (1988). Cognitive load during problem solving: Effects on
learning. \textit{Cognitive Science}, \textit{12}(2), 257--285.

Wu, X., et al. (2024). LongMemEval: Benchmarking chat assistants on
long-term interactive memory.
\url{https://doi.org/10.48550/arXiv.2410.10813}

Xu, W., et al. (2025a). A-MEM: Agentic memory for LLM agents.
\url{https://doi.org/10.48550/arXiv.2502.12110}

Xu, W., et al. (2025b). Emotional RAG: Enhancing role-playing agents
through emotional retrieval.
\url{https://doi.org/10.48550/arXiv.2410.23041}

Zhang, J., et al. (2024a). EmotionQueen: A benchmark for evaluating
empathy of large language models.
\url{https://doi.org/10.48550/arXiv.2409.13359}

Zhang, J., et al. (2024b). Enabling personalized long-term interactions
in LLM-based agents through persistent memory and user profiles.
\url{https://doi.org/10.48550/arXiv.2510.07925}

Zhang, J., et al. (2025). Personalizing emotion-aware conversational
agents.
\url{https://doi.org/10.48550/arXiv.2511.06954}

Zheng, L., et al. (2023). Judging LLM-as-a-judge with MT-Bench and
Chatbot Arena. \textit{NeurIPS 2023}.
\url{https://arxiv.org/abs/2306.05685}

Zhong, W., et al. (2024). MemoryBank: Enhancing large language models
with long-term memory. \textit{AAAI 2024}.
\href{https://arxiv.org/abs/2305.10250}{arXiv:2305.10250}.

\endgroup

\end{document}